\pdfoutput=1

\documentclass[11pt]{article}

\usepackage{acl}
\usepackage{nert}

\usepackage{times}
\usepackage{latexsym}
\usepackage{tabularx}

\usepackage[utf8]{inputenc}

\title{\textsc{BabyReasoningBench}: Generating Developmentally-Inspired Reasoning Tasks for Evaluating Baby Language Models}

\author{Kaustubh D. Dhole \\ 
  Department of Computer Science \\
  Emory University \\
  \eml{kdhole@emory.edu}
}
\date{}

\begin{document}
\maketitle

\begin{abstract}
Traditional evaluations of reasoning capabilities of language models are dominated by adult-centric benchmarks that presuppose broad world knowledge, complex instruction following, and mature pragmatic competence. These assumptions are mismatched to~\emph{baby language models} trained on developmentally plausible input such as child-directed speech and early-childhood narratives, and they obscure which reasoning abilities (if any) emerge under such constraints. We introduce~\textbf{~\textsc{BabyReasoningBench}}\footnote{\url{https://github.com/kaustubhdhole/baby-reasoning-bench}}, a GPT-5.2 generated benchmark of 19 reasoning tasks grounded in classic paradigms from developmental psychology, spanning theory of mind, analogical and relational reasoning, causal inference and intervention selection, and core reasoning primitives that are known to be confounded by memory and pragmatics. We find that two GPT-2 based baby language models (pretrained on 10M and 100M of child-directed speech text) show overall low but uneven performance, with dissociations across task families: scaling improves several causal and physical reasoning tasks, while belief attribution and pragmatics-sensitive tasks remain challenging.~\textsc{BabyReasoningBench} provides a developmentally grounded lens for analyzing what kinds of reasoning are supported by child-like training distributions, and for testing mechanistic hypotheses about how such abilities emerge.\end{abstract}

\section{Introduction}

Large language models (LLMs) are typically evaluated on adult-centric benchmarks that presume extensive world knowledge, long-form instruction following, and mature linguistic competence~\cite{phan2025humanity,srivastava2023beyond,dhole2025conqret}. This evaluation regime makes it difficult to answer a different question that matters for both cognitive modeling and data-efficient AI~\cite{ghanizadeh-dousti-2024-towards}: \emph{what kinds of reasoning emerge when models are trained primarily on developmentally plausible input}? In particular, ``baby language models''---models trained on caregiver--child interaction data~\cite{warstadt2023findings,feng2024child}, early-childhood text, and simplified perceptual or narrative inputs---invite evaluation paradigms aligned with the \emph{developmental trajectory} of human cognition rather than expert adult performance. They also enable causal experiments (e.g., controlled-rearing studies) that can be hard to perform on children~\cite{rozner-etal-2025-babylms}.

We introduce \textbf{~\textsc{BabyReasoningBench}}, a curated collection of tasks grounded in classic findings from developmental psychology and infant cognition. The collection spans (i) \emph{theory-of-mind} and belief attribution via explicit false-belief elicitation in preschoolers \cite{WimmerPerner1983,BaronCohenLeslieFrith1985} as well as implicit violation-of-expectation paradigms in infants \cite{OnishiBaillargeon2005} and broader batteries summarized meta-analytically \cite{WellmanCrossWatson2001}; (ii) \emph{analogical reasoning} and relational mapping across familiar causal transformations and story structures \cite{GoswamiBrown1990,GentnerToupin1986,HolyoakJunnBillman1984}; (iii) \emph{causal learning} from evidence patterns (e.g., ``blicket detector'' inferences) \cite{GopnikSobel2000}, causal-structure induction and Bayesian-inspired learning accounts \cite{GopnikEtAl2004}, and curiosity-driven exploration under confounding \cite{SchulzBonawitz2007}; and (iv) \emph{core reasoning primitives} that are known to be sensitive to memory, pragmatics, and inhibitory control, including transitive inference \cite{BryantTrabasso1971}, counterfactual and counterfactual-as-possibility reasoning \cite{HarrisGermanMills1996,BeckEtAl2006}, category-based induction \cite{GelmanMarkman1986}, pragmatic wording effects in class-inclusion queries \cite{Shipley1979}, scientific control-of-variables reasoning \cite{ChenKlahr1999}, and conservation under accidental versus intentional transformations \cite{McGarrigleDonaldson1974}.

Evaluating baby language models on these paradigms serves three complementary goals. First, it anchors model behavior to \emph{human developmental baselines}: many tasks exhibit sharp age-related transitions (e.g., false-belief performance shifting between ages 3--5 \cite{WellmanCrossWatson2001}), while others reveal competence under reduced language demands (e.g., implicit false-belief expectations in 15-month-olds \cite{OnishiBaillargeon2005}). Second, developmental tasks often disentangle reasoning from confounds such as pragmatic interpretation \cite{Shipley1979,McGarrigleDonaldson1974} or executive demands \cite{BryantTrabasso1971,BeckEtAl2006}, providing diagnostic leverage beyond aggregate accuracy. Third, these paradigms encourage \emph{mechanism-sensitive} evaluation: success can depend on representing others' beliefs, mapping relational structure, integrating evidence across trials, or proposing informative interventions---abilities that may (or may not) emerge under developmentally realistic training distributions.

Table~\ref{tab:BabyReasoningBench_tasks} summarizes the task families in ~\textsc{BabyReasoningBench} and the corresponding empirical signatures in human infants and children that motivate each evaluation setting.

\section{~\textsc{BabyReasoningBench}}
Many of the reasoning tasks traditionally evaluated on children are straightforward for frontier models\footnote{While many of these traditional tasks are not necessarily evaluated through verbal cues, we verbalize each task as MCQs}, and several of them (e.g., the Sally--Anne test) have been extensively discussed across several books and online making LLM based generation useful for such tasks. To construct~\textsc{~\textsc{BabyReasoningBench}} in a way that is systematic and less sensitive to idiosyncratic question phrasing, we proceed as follows. We first provide OpenAI GPT-5.2~\footnote{\url{https://openai.com/index/introducing-gpt-5-2/}} with a template containing the 19 tasks discussed in this paper, including a brief description of each task in tabular form, and instruct it to generate one multiple-choice question per task. After manually validating these seed questions, we then prompt GPT-5.2 to produce 10 additional variants for each seed, yielding 11 questions per task in total. This replication strategy helps normalize against potential pattern biases in baby language models by reducing reliance on any single wording or surface form. Finally, we manually review all generated MCQs and apply minor edits where questions or answers appear unreasonable, like missing context. Wherever possible, we try to keep the token length of the the answer choices similar, so our perplexity-based multiple-choice classification is less confounded by token counts, but we do not employ it as a hard constraint.

As a preliminary evaluation, we evaluate two BabyLM baselines trained on developmentally motivated corpora: BabyLM-10M~\footnote{\url{https://hf.co/BabyLM-community/babylm-baseline-10m-gpt2}\relax} and BabyLM-100M~\footnote{\url{https://hf.co/BabyLM-community/babylm-baseline-100m-gpt2}\relax}~\cite{charpentier2025babylmturns3papers} as both models have beene been pretrained on several datasets, including CHILDES~\cite{macwhinney2014childes}, using a next-token prediction objective. For each question, we build a prompt for each choice by concatenating the question and the choice, then score each choice by the conditional log-likelihood of its tokens given the question under the language model. The predicted answer is chosen as the option with the highest summed log-probability.

\begin{figure*}[t]
    \centering
    \includegraphics[width=.85\linewidth]{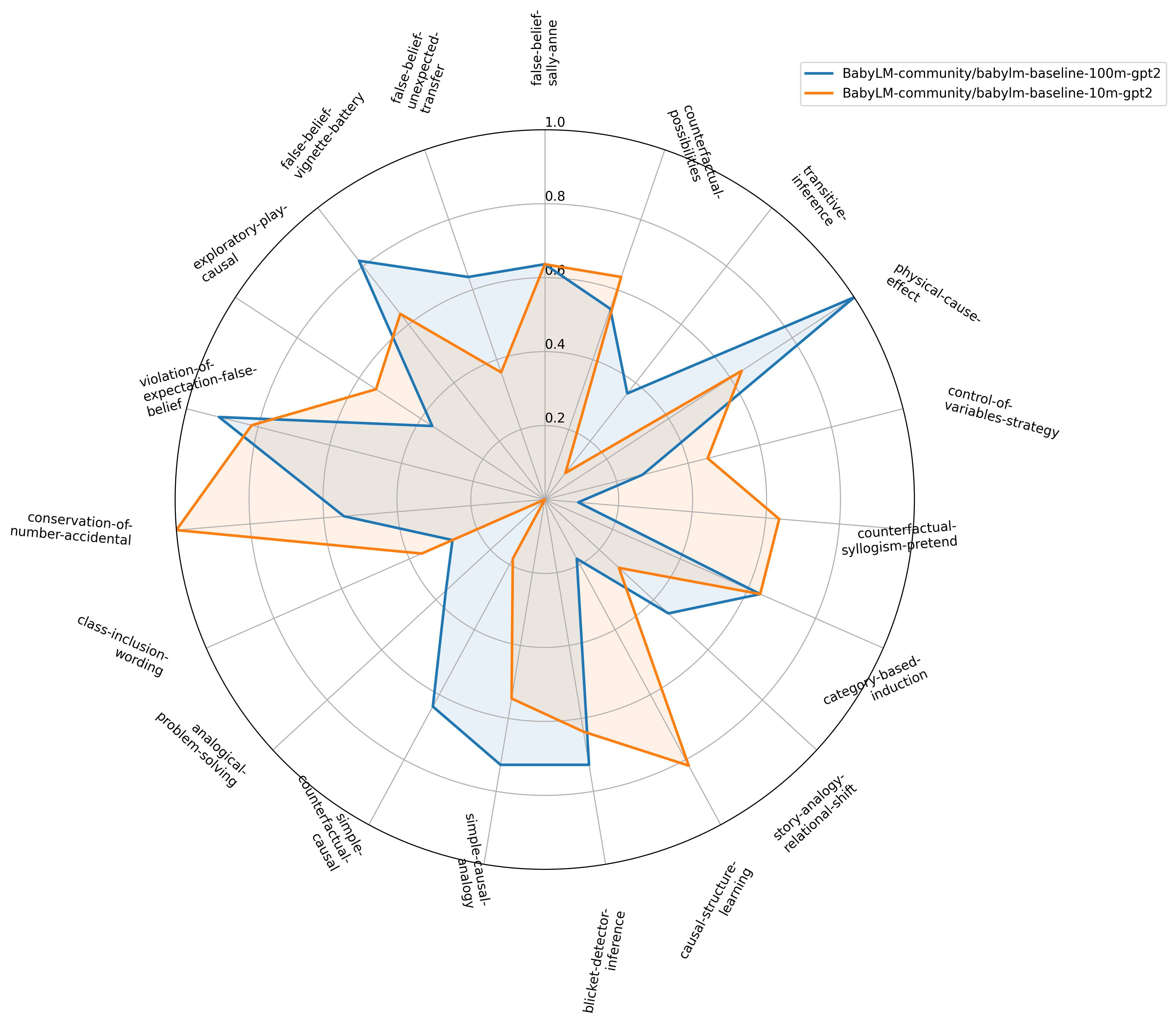}
    \caption{Performance of a 10M and 100M GPT2 evaluated on~\textsc{BabyReasoningBench}}
    \label{fig:spider}
\end{figure*}


\begin{table}[t]
\centering
\small
\setlength{\tabcolsep}{4pt}
\begin{tabularx}{\columnwidth}{@{}Xrr@{}}
\hline
\textbf{Task} & \textbf{10M} & \textbf{100M} \\
\hline
False Belief Sally Anne                 & 63.64 & 63.64 \\
Counterfactual Possibilities            & 63.64 & 54.55 \\
Transitive Inference                    & 9.09  & 36.36 \\
Physical Cause Effect                   & 63.64 & 100.00 \\
Control of Variables Strategy           & 45.45 & 27.27 \\
Counterfactual Syllogism Pretend        & 63.64 & 9.09 \\
Category Based Induction                & 63.64 & 63.64 \\
Story Analogy Relational Shift          & 27.27 & 45.45 \\
Causal Structure Learning               & 81.82 & 18.18 \\
Blicket Detector Inference              & 63.64 & 72.73 \\
Simple Causal Analogy                   & 54.55 & 72.73 \\
Simple Counterfactual Causal            & 18.18 & 63.64 \\
Analogical Problem Solving              & 0.00  & 36.36 \\
Class Inclusion Wording                 & 36.36 & 27.27 \\
Conservation of Number Accidental       & 100.00 & 54.55 \\
Violation of Expectation False Belief   & 81.82 & 90.91 \\
Exploratory Play Causal                 & 54.55 & 36.36 \\
False Belief Vignette Battery           & 63.64 & 81.82 \\
False Belief Unexpected Transfer        & 36.36 & 63.64 \\
\hline
\end{tabularx}
\caption{Task-wise accuracy for two BabyLM GPT-2 baselines (pretrained on 10M and 100M of child-directed speech respectively).}
\label{tab:babylm_two_models_tasks}
\end{table}

\begin{table*}[h]
\centering
\small
\begin{tabular}{p{0.28\linewidth} p{0.36\linewidth} p{0.30\linewidth}}
\toprule
\textbf{Reasoning Task}  & \textbf{Description} & \textbf{Performance of Children} \\
\midrule
False-belief (unexpected transfer) \cite{WimmerPerner1983}
& Classic location-change vignette: predict a person's search given the person did not observe a move.
& Children under $\sim$4 tend to answer using reality; older preschoolers succeed. \\[2pt]

False-belief (Sally--Anne; autism contrast) \cite{BaronCohenLeslieFrith1985}
& Puppet/doll false-belief story; 
& Typical 4-year-olds frequently pass; autistic children show markedly lower pass rates. \\[2pt]

False-belief battery (meta-analytic factors) \cite{WellmanCrossWatson2001}
& Multiple variants (unexpected contents, transfer, wording/motivation manipulations).
& Strong age effect from 3$\rightarrow$5 years. \\[2pt]

Implicit false-belief (violation-of-expectation) \cite{OnishiBaillargeon2005}
& Nonverbal expectation measure: infants look longer when an actor acts inconsistently with her belief.
& Evidence for belief-consistent expectations in 15-month-olds under reduced language demands. \\[2pt]

Causal analogies with familiar relations \cite{GoswamiBrown1990}
& Four-term analogies based on understood causal transformations (e.g., melting).
& Even 3--4-year-olds can succeed when relations are familiar/causally transparent. \\[2pt]

Relational mapping vs.\ surface similarity (story analogy) \cite{GentnerToupin1986}
& Re-enact a base story with new characters; vary surface similarity and relational clarity.
& Older children show ``relational shift'' (prefer deep structure); younger children rely more on surface similarity. \\[2pt]

Analogical problem solving / transfer with support \cite{HolyoakJunnBillman1984}
& Hear a base story/pattern; solve a new problem that shares structure; hints may be provided.
& Preschoolers can transfer with scaffolding and close mappings; older children generalize more flexibly. \\[2pt]

Physical cause--effect principles \cite{BullockGelmanBaillargeon1982}
& Judgments about whether effects can occur without causes; temporal priority (cause-before-effect).
& Preschoolers show early understanding that causes precede effects and that ``magic'' effects are dispreferred. \\[2pt]

Blicket detector causal inference \cite{GopnikSobel2000}
& Infer which objects have hidden causal power from activation patterns across trials.
& 2--4-year-olds integrate evidence across trials to identify causal candidates. \\[2pt]

Causal structure learning (Bayes nets account) \cite{GopnikEtAl2004}
& Infer latent causal graphs from observations/interventions; predict outcomes of new interventions.
& Empirical and theoretical account of strong causal learning abilities emerging around ages 2--4. \\[2pt]

Exploratory play under confounding \cite{SchulzBonawitz2007}
& Free play after confounded vs.\ unconfounded evidence; measure information-seeking behaviors.
& Preschoolers explore more when evidence is ambiguous/confounded; some spontaneously isolate variables. \\[2pt]

Transitive inference (memory demands) \cite{BryantTrabasso1971}
& Learn ordered relations (A$>$B, B$>$C); infer A$>$C, especially with memory supports.
& Young children can succeed when memory demands are reduced. \\[2pt]

Counterfactual syllogisms in pretend contexts \cite{DiasHarris1988}
& Deduction from counterfactual premises (``All bears fly...''); compare standard vs.\ make-believe framing.
& Preschoolers more often follow logic when explicitly invited to pretend rather than defaulting to real-world knowledge. \\[2pt]

Category-based induction over appearance \cite{GelmanMarkman1986}
& Project a novel property to a target: category match vs.\ perceptual similarity conflict.
& By around age 4, children often privilege category membership over surface similarity. \\[2pt]

Class inclusion / wording effects \cite{Shipley1979}
& Part--whole comparisons (``more dogs or more animals?''); manipulate question form and distributive vs.\ collective readings.
& Performance improves substantially with pragmatically clearer wording that matches ordinary English usage. \\[2pt]

Causal counterfactual ``what-if'' \cite{HarrisGermanMills1996}
& Simple causal narratives; ask what would happen if a key cause were altered.
& 3--4-year-olds answer simple causal counterfactuals correctly \\[2pt]

Counterfactuals as alternative possibilities \cite{BeckEtAl2006}
& Compare answering a single counterfactual vs.\ representing multiple possible alternatives.
& Younger children often revert to reality; more consistent ``possibility'' reasoning emerges later \\[2pt]

Control of Variables Strategy (CVS) \cite{ChenKlahr1999}
& Multi-factor experiments; test whether the child varies one variable at a time and transfers the strategy.
& 8--10-year-olds acquire and transfer CVS only with training \\[2pt]

Conservation with accidental vs.\ intentional transformation \cite{McGarrigleDonaldson1974}
& Conservation of number under ``naughty teddy'' accidental change vs.\ adult intentional change.
& Many more 4--6-year-olds conserve under accidental transformations, implicating pragmatic/intent interpretations. \\
\bottomrule
\end{tabular}
\caption{~\textsc{BabyReasoningBench} tasks: developmental paradigms that motivate evaluation of baby language models, with performance patterns of children studied in human behavioral tests.}
\label{tab:BabyReasoningBench_tasks}
\end{table*}

\section{Results}
Table~\ref{tab:babylm_two_models_tasks} and Figure~\ref{fig:spider} report task-wise accuracy for two BabyLM GPT-2 baselines. We find moderate to strong overall performance for both models, with only a modest difference between scales. The average accuracy is \textbf{52.15\%} for the 10M model and \textbf{53.59\%} for the 100M model, indicating that both models solve a meaningful subset of the benchmark, but in a highly non-uniform way across paradigms.
\\
\paragraph{Scaling yields selective gains, especially on some physical, causal, and analogy tasks.}
The 100M model outperforms the 10M model on several tasks that involve local causal structure or relational transfer. It reaches \textbf{100.00\%} on \texttt{physical-cause-effect} (vs.\ 63.64\% at 10M), improves on \texttt{blicket-detector-inference} (63.64$\rightarrow$72.73), \texttt{simple-causal-analogy} (54.55$\rightarrow$72.73), \texttt{simple-counterfactual-causal} (18.18$\rightarrow$63.64), and \texttt{analogical-problem-solving} (0.00$\rightarrow$36.36), and also performs strongly on \texttt{violation-of-expectation-false-belief} (90.91\%). These gains suggest that additional pre-training can help on tasks that require short-range evidence integration or belief-consistent prediction.
\\
\paragraph{At the same time, the benchmark does not show monotonic scaling.}
The 10M model matches or exceeds the 100M model on a number of tasks, including \texttt{counterfactual-possibilities}, \texttt{control-of-variables-strategy}, \texttt{counterfactual-syllogism-pretend}, \texttt{causal-structure-learning}, \texttt{class-inclusion-wording}, \texttt{conservation-of-number-accidental}, and \texttt{exploratory-play-causal}. The largest reversal appears in \texttt{causal-structure-learning}, where the 10M model scores 81.82\% and the 100M model only 18.18\%. This pattern indicates that developmentally grounded reasoning performance is not well captured by a simple ``larger is better'' account, and may depend strongly on task framing and model-specific heuristics.
\\
\paragraph{Belief attribution and pragmatics-sensitive reasoning are fragile rather than absent.}
The rerun results also revise the interpretation of theory-of-mind-related tasks. Both models are now clearly above floor on explicit false-belief paradigms: \texttt{false-belief-sally-anne} is 63.64\% for both models, \texttt{false-belief-vignette-battery} reaches 63.64\% and 81.82\%, and \texttt{false-belief-unexpected-transfer} reaches 36.36\% and 63.64\%. Likewise, \texttt{class-inclusion-wording} is no longer at floor, though performance remains modest. This suggests that these models can exhibit some belief-sensitive and pragmatics-sensitive behavior, but that such competence is unstable across formats rather than robustly generalized.

Overall, ~\textsc{BabyReasoningBench} continues to reveal strong dissociations across developmental reasoning paradigms, but the updated results point to a more nuanced conclusion: both models show broader competence but overall still moderate performance, while scaling yields only selective and inconsistent benefits. This reinforces the value of fine-grained developmental evaluation for distinguishing robust reasoning abilities from brittle task-specific successes.
\section{Conclusion}

We introduced \textbf{\textsc{BabyReasoningBench}}, a benchmark of simple developmental reasoning tasks designed to evaluate \emph{baby language models} inspired by classic experimental paradigms from developmental psychology. Because most of these tasks, when framed textually, are trivial for modern frontier models, they are especially useful as a diagnostic benchmark for small, developmentally trained models. The benchmark operationalizes theory of mind, analogical reasoning, causal inference, and several core reasoning primitives through multiple-choice questions, with task variants generated using an LLM (GPT-5.2) to systematically vary surface form while preserving logic.

Baseline results on two BabyLM GPT-2 models show that (i) performance is substantially below ceiling overall, (ii) increases in pre-training data yield meaningful gains on multiple causal and physical reasoning tasks, but (iii) belief attribution and pragmatics-sensitive tasks remain fragile and highly format-dependent rather than uniformly absent. The task-level dissociations—including sharp failures on class inclusion wording and inconsistent outcomes across false-belief formats—underscore why fine-grained developmentally grounded evaluation is valuable: it can separate reasoning competence from confounds tied to language form, framing, and inhibitory demands, and it can expose brittle generalization that would be invisible in aggregate benchmarks.

~\textsc{BabyReasoningBench} is intended as a diagnostic instrument rather than a leaderboard-only dataset. Future work would (a) expand the tasks with other reasoning capabilities, (b) incorporate controlled manipulations of wording, memory load, and discourse context to disentangle pragmatics from reasoning, and (c) extend the suite to multimodal and interactive settings (e.g., simplified visual scenes and intervention selection) that more closely match the evidence available to infants and young children in behavioral experiments. We hope this benchmark supports more mechanism-sensitive evaluation of developmentally trained models and fine-grained empirical evaluations of which reasoning abilities can emerge from child-like input distributions.

\section{Limitations}
Our work has some limitations, primarily stemming from the verbalisation of the tasks. While some of the tasks, like evaluating for analogies and counterfactuals, maintain their essence in textual form, others, like the Sally-Anne task and the blicket task, may arguably be better assessed with multimodal input, like their behavioral counterparts. 

For some tasks, the MCQs may represent a subset of the actual reasoning behavior. For instance, the blinket task evaluates causal inference using indirect evidence but not disjunctive or conjunctive inference. We leave further fine-grained analysis, and additional MCQs for subsequent work. 

\section{Ethical Consideration}
The paper has been written with the help of OpenAI ChatGPT 5.2, and Grammarly. The ideas in the paper are solely from the author.

\section*{Acknowledgements}
The author thanks Michael C. Frank (Department of Psychology, Stanford University) for helpful initial discussions.

\bibliography{mypaper}


\end{document}